# Machine learning prediction of obstructive coronary artery disease using opportunistic coronary calcium and epicardial fat assessments from CT calcium scoring scans

**Juhwan Lee[1,2,*], Ammar Hoori[2], Tao Hu[2], Justin N. Kim[2], Mohamed H. E. Makhlouf[3], Michelle C. Williams[4], David E. Newby[4], Robert Gilkeson[3], Sanjay Rajagopalan[3], David L. Wilson[2,5]**

[1] *Department of Biomedical Engineering, Virginia Commonwealth University, Richmond, VA, 23284, USA*
[2] *Department of Biomedical Engineering, Case Western Reserve University, Cleveland, OH, 44106, USA*
[3] *Harrington Heart and Vascular Institute, University Hospitals Cleveland Medical Center, Cleveland, OH, 44106, USA*
[4] *BHF Centre for Cardiovascular Science, University of Edinburgh, Edinburgh, UK*
[5] *Department of Radiology, Case Western Reserve University, Cleveland, OH, 44106, USA*

*Corresponding author: jxl1982@case.edu
Telephone number: 216-258-3527

## Abstract

Non-contrast computed tomography calcium scoring (CTCS) is a cost-effective imaging modality widely used to detect coronary artery calcifications. This study aimed to develop an advanced machine learning framework that utilizes quantitative analyses of coronary calcium and epicardial fat from CTCS images to predict obstructive coronary artery disease (CAD). The study population consisted of 1,324 patients from the SCOT-HEART clinical trial who underwent both CTCS and coronary CT angiography. We extracted and analyzed a broad range of features, including 24 clinical variables, 189 calcium-omics, and 211 epicardial fat-omics features from the CTCS images. Feature selection was conducted using the CatBoost algorithm combined with SHapley Additive exPlanation (SHAP) values. Predictive modeling utilized the CatBoost gradient boosting method, focusing on the most informative features. From an initial set of 424 candidate features, 14 were identified as most predictive through the CatBoost-SHAP method. The top two predictive features originated from fat-omics, with the remaining 12 features derived from calcium-omics. The optimized model achieved robust predictive capabilities, demonstrating a sensitivity of 83.1±4.6%, specificity of 93.8±1.7%, accuracy of 85.3±2.0%, and an F1 score of 73.9±3.3%. Inclusion of calcium-omics and fat-omics data significantly improved predictive performance. Notably, the model also showed reliable predictive accuracy in patients with diverse coronary calcium scores, including cases with obstructive CAD despite a zero-calcium score. This innovative approach holds promise for improving clinical decision-making and potentially reducing dependence on contrast-enhanced or invasive diagnostic procedures, particularly within low-to intermediate-risk patient groups.

**Keywords**: Obstructive coronary artery disease, computed tomography calcium scoring, calcium-omics, fat-omics, machine learning, classification

# 1 Introduction

Coronary computed tomography angiography (CCTA) has emerged as an important imaging modality for evaluating obstructive coronary artery disease (CAD). By providing detailed, high-resolution visualization of the coronary arteries, CCTA significantly improves clinicians' ability to accurately diagnose and manage CAD. To facilitate standardized interpretation of CCTA images, the Coronary Artery Disease Reporting and Data System (CAD-RADS) was developed. This classification ranges from CAD-RADS 0, representing the absence of plaque or stenosis, to CAD-RADS 5, indicating complete arterial occlusion, along with additional categories for patients with prior coronary interventions [1]. Numerous studies and clinical trials have validated the CAD-RADS method, highlighting its reliability for clinical decision-making and patient risk stratification [2–4]. Nevertheless, routine clinical adoption of CCTA faces several limitations, including higher costs, reliance on iodine-based contrast agents, inter-reader variability, and concerns regarding ionizing radiation exposure.

In contrast, non-contrast CT calcium scoring (CTCS) is a cost-effective imaging technique that directly detects coronary artery calcifications, an established marker of coronary atherosclerosis. Recognized by multiple clinical guidelines, CTCS is particularly valued for its affordability, simplicity, and substantially lower radiation exposure compared to contrast-enhanced CCTA [5,6]. Despite its well-established role in cardiovascular risk prediction, the use of CTCS specifically for diagnosing obstructive CAD remains less explored, creating an opportunity to improve its diagnostic utility. Previously, our research demonstrated that an automated machine learning approach leveraging quantitative analysis of epicardial adipose tissue in CTCS images could reliably predict obstructive CAD across diverse multi-center populations [7]. Building upon these findings [7], the current study further hypothesized that quantitative assessments of coronary calcification and epicardial adipose tissue from CTCS scans will correlate significantly with obstructive CAD as determined by CCTA. Given the unique insights into plaque burden provided by CTCS, exploring these quantitative features could notably improve cardiovascular risk stratification.

In this study, we developed a novel predictive framework utilizing machine learning algorithms designed to identify obstructive CAD from readily accessible, non-contrast CTCS scans. To achieve this, we evaluated multiple clinical characteristics, calcium-omics biomarkers [8], and epicardial fat-omics features [9] to identify the most robust predictors. Following feature selection, we systematically trained and compared several advanced machine learning models to evaluate their diagnostic performance. Our findings are anticipated to improve clinical decision-making processes and facilitate more effective cardiovascular risk assessment through widely available, cost-effective imaging.

# 2 Methods

*2.1 Study cohort*

This study was conducted as a secondary analysis of the Scottish COmputed Tomography of the HEART (SCOT-HEART) multicenter randomized controlled trial (ClinicalTrials.gov identifier: NCT01149590). Ethical approval for the original trial was obtained from the respective local ethics committees, and all enrolled participants provided written informed consent. The current analysis was performed in accordance with a data use agreement established between the University of Edinburgh (Edinburgh, UK) and University Hospitals Cleveland Medical Center (Cleveland, Ohio, USA). Briefly, of the initial 4,146 patients referred to cardiology outpatient clinics in the trial, 2,073 were assigned to the intervention arm, among whom 1,778 underwent CT scanning. Of these, 1,324 CT scans met quality standards and were thus eligible for inclusion in the present analysis. Further details and primary results from the trial have been reported previously [10,11].

*2.2 CT image acquisition*

All study participants underwent two imaging protocols: electrocardiogram-gated CTCS without contrast and electrocardiogram-gated CCTA with iodine-based contrast enhancement. Imaging acquisitions were performed utilizing CT scanners equipped with either 64 or 320 detector rows, specifically the Brilliance 64 (Philips Medical Systems, Netherlands), Biograph mCT (Siemens, Germany), or Aquilion ONE (Toshiba Medical Systems, Japan). Scanner parameters, including tube voltage, tube current, and the volume of iodine-based contrast medium, were individually tailored according to each participant's body mass index (BMI). Standard electrocardiogram gating techniques were applied to synchronize image acquisition with cardiac phases, ensuring accurate depiction of coronary anatomy.

*2.3 Identification and classification of obstructive CAD based on CAD-RADS*

All CCTA scans underwent CAD-RADS assessment using standardized semi-automatic analysis software (AutoPlaque, Version 2.5, Cedars-Sinai Medical Center, Los Angeles, California, USA). The CAD-RADS classification systematically categorized CAD severity into distinct groups based on the maximal stenosis percentage identified in any coronary vessel. Specifically, the classification scheme included the following categories: CAD-RADS 0 (0% stenosis, indicating no detectable CAD), CAD-RADS 1 (1-24% stenosis, minimal non-obstructive CAD), CAD-RADS 2 (25-49% stenosis, mild non-obstructive CAD), CAD-RADS 3 (50-69% stenosis, moderate stenosis), CAD-RADS 4A (70-99% stenosis, severe stenosis in one or two vessels), CAD-RADS 4B (> 50% stenosis involving the left main artery or ≥70% stenosis affecting three vessels), and CAD-RADS 5 (100% stenosis, representing complete vessel occlusion).

In the current analysis, obstructive CAD was defined by a CAD-RADS classification of 4A, 4B, or 5, whereas categories 0 through 3 were classified as non-obstructive CAD. Each scan was independently reviewed, and final CAD-RADS categories were confirmed by experienced imaging specialists to ensure accuracy and consistency.

*2.4 Feature extraction and analysis*

To predict obstructive CAD from CTCS scans, a comprehensive set of clinical and imaging-based features was extracted and analyzed. Specifically, we examined a total of 424 features, categorized as 24 clinical, 189 calcium-omics, and 211 fat-omics features. Each feature group was carefully curated to capture relevant biological and clinical information associated with obstructive CAD.

**Clinical features** included demographic and baseline patient characteristics (e.g., age, gender, and BMI), cardiovascular risk factors, laboratory blood test results (e.g., total cholesterol, LDL cholesterol, and HDL cholesterol), and medication prescriptions (e.g., statins, angiotensin-converting enzyme [ACE] inhibitors, and beta-blockers). Additionally, chest pain symptoms were systematically classified into cardiac and non-cardiac chest pain categories based on established clinical guidelines and recommendations [12,13].

**Calcium-omics features** comprise 189 quantitative imaging biomarkers systematically extracted at multiple spatial scales, including individual calcified lesions, specific coronary artery territories, and the entire cardiac region (Fig. 1). At the level of individual calcifications, we extracted elemental radiomic features such as calcium mass, lesion volume, Hounsfield Unit (HU) statistics (minimum, maximum, and mean values), first- and second-order statistical moments, lesion shape descriptors, and spatial metrics including the distance to subsequent calcified lesions, distance from the lesion to the top of the CT volume, and the anatomical territory location. Additionally, we quantified artery diffusivity to characterize the spatial distribution and dispersion of calcified lesions with each coronary artery territory. Diffusivity was set to 0 for territories with no lesions and to 1 for territories containing only a single calcification. At the coronary territory and whole-heart levels, we computed aggregate statistical measures, including mean, standard deviation, skewness, kurtosis, and histogram-based metrics to comprehensively capture the distribution of calcification burden across the coronary arteries and the overall cardiac volume. Conventional clinical metrics such as Agatston score, calcium mass, and lesion volume were calculated at each scale – individual lesion, artery territory, and whole heart – to facilitate both detailed and broad assessments of calcified plaque burden. Detailed methodology for calcium-omics feature extraction has been previously described and validated elsewhere [8].

**Fat-omics features** consisted of 211 handcrafted imaging biomarkers inspired by pathophysiological processes, systematically grouped into morphological, intensity-based, and spatial distribution categories (Fig. 1) [9]. To obtain these fat-omics biomarkers, epicardial adipose tissue regions enclosed within the pericardium were segmented using our previously validated automated segmentation network, DeepFat [14]. Briefly, the pericardial boundaries were automatically delineated using a DeepLab V3+ network, and epicardial fat was subsequently defined as the adipose tissue located within the segmented pericardium and characterized by HU values ranging from −190 to −30. After segmentation, detailed fat-omics features were extracted. Morphological features captured structural attributes of EAT, including total volume, principal axis lengths, epicardial fat thickness, and surface area. Intensity-based features encompassed statistical descriptors of adipose tissue attenuation, such as minimum, maximum, mean HU values, skewness, kurtosis, entropy, and histogram-based metrics. To quantitatively assess spatial fat distribution, the segmented cardiac volume was systematically partitioned into four equal-thickness axial slabs from superior to inferior and radially subdivided into four equidistant concentric ribbons extending from the outer pericardial boundary toward the heart center. This approach allowed comprehensive spatial characterization of EAT distribution patterns within the pericardium. Detailed descriptions of these fat-omics feature extraction methods have been reported previously [9].

*2.5 Feature selection*

We performed feature selection using the CatBoost algorithm [15] in combination with the SHapley Additive exPlanations (SHAP) method [16]. The CatBoost model was trained using the following optimized hyperparameters: 300 iterations, a learning rate of 0.01, a tree depth of 6, L2 leaf regularization set to 5, random subspace method of 0.75, border count of 64, Bernoulli bootstrapping with subsample rate of 0.6, and automatic class weighting to handle class imbalance. The binary logistic loss function was used for both training and evaluation, and computations were parallelized using 10 computational threads. Additionally, early stopping was applied, retaining the iteration with the best validation metric. To prevent overfitting and improve the robustness of the model, we utilized 5-fold cross-validation, wherein the dataset was randomly partitioned into five subsets. The CatBoost model was iteratively trained on four subsets and validated on the remaining subset. This approach ensured the generalizability of the model to unseen data. All feature selection and subsequent model training procedures were strictly performed using only the training folds. Finally, to rank and select the most relevant features, we applied the SHAP approach, calculating feature importance based on the average absolute SHAP values derived from the cross-validated CatBoost models.

*2.6 Development of machine learning models*

We developed predictive models using the CatBoost gradient boosting algorithm [15], trained on the most important features identified through the CatBoost-SHAP feature selection method described above. CatBoost was selected for its robustness in handling heterogeneous medical datasets and its effectiveness in capturing complex feature interactions. Model optimization involved an extensive hyperparameter tuning using a grid search approach, evaluating a broad range of parameters, including the number of iterations, learning rate, depth, L2 leaf regularization, random subspace method, border count, loss function, evaluation metric, bootstrap type, subsample, and thread count. Supplementary Table S1 provides detailed hyperparameter search ranges and the optimal hyperparameter values identified through this tuning process for each specific model.

To systematically evaluate the incremental predictive value of imaging-derived features, we trained and assessed three separate machine learning models with different feature combinations: ***Model 1*** included clinical features only; ***Model 2*** combined clinical features with calcium-omics; and ***Model 3*** integrated clinical, calcium-omics, and fat-omics features. All three models underwent identical cross-validation and hyperparameter tuning procedures to ensure comparability and robustness of results.

*2.7 Performance evaluation*

To rigorously evaluate the robustness and reliability of our predictive models, we performed 1,000 iterations of 5-fold cross-validation. In each iteration, the dataset was partitioned into five subsets, training was conducted on four subsets, and the remaining subset was reserved for validation. This repeated cross-validation procedure ensured comprehensive assessment of model stability, reduced variability, and mitigated potential overfitting.

The classification performance of the models was quantitatively measured using standard evaluation metrics, including sensitivity, specificity, accuracy, the area under the receiver operating characteristics curve (AUROC), and the area under the precision-recall curve (AUPRC). These metrics were calculated as follows:

$$Sensitivity = TP / (TP + FN) \quad (1)$$

$$Specificity = TN / (TN + FP) \quad (2)$$

$$\text{Accuracy} = (TP + TN) / (TP + FP + TN + FN) \quad (3)$$

where TP denotes true positives, TN represents true negatives, FP indicates false positives, and FN refers to false negatives. We reported the mean and standard deviation of each metric across the 1,000 iterations to provide an accurate estimate of model performance and its variance.

To ensure a thorough evaluation of our optimized CatBoost model, we compared its predictive performance against three alternative machine learning algorithms commonly used in classification tasks: Support Vector Machine (SVM), Random Forest (RF), and XGBoost. All comparison models underwent identical cross-validation procedures and were trained using the same dataset partitions to ensure fair and direct performance comparisons.

*2.8 Statistical analysis*

Continuous variables were reported as mean ± standard deviation, and categorical variables were summarized as frequencies and percentages. The normality of continuous data was assessed using the Shapiro-Wilk test. Comparisons between groups (obstructive vs. non-obstructive CAD) were performed using Student's t-test for normally distributed continuous variables and the Chi-square test for categorical variables. Statistical significance was defined as a two-tailed p-value less than 0.05. To evaluate differences in predictive performance among models, McNemar's test was used to compare sensitivities and specificities, and the DeLong test was applied to compare the AUROC and AUPRC values. All statistical analyses were conducted using R Studio software (version 2024.04.1, R Foundation for Statistical Computing, Vienna, Austria).

## 3 Results

In our study cohort of 1,324 patients, obstructive CAD was present in 334 (25.2%). Patients were distributed across CAD-RADS categories as follows: CAD-RADS 0 in 472 patients (35.6%), CAD-RADS 1 in 248 (18.7%), CAD-RADS 2 in 150 (11.3%), CAD-RADS 3 in 120 (9.1%), CAD-RADS 4A in 160 (12.1%), CAD-RADS 4B in 30 (2.3%), and CAD-RADS 5 in 144 (10.9%). Patients with obstructive CAD were significantly older, with a mean age of 61.2±7.7 years, compared to 55.9±9.5 years for patients without obstructive CAD ($p<0.00001$). Male participants comprised 76% of the obstructive CAD group and 48.8% of the non-obstructive CAD group. However, this difference was not statistically significant ($p>0.05$). While total cholesterol levels did not significantly differ between groups, HDL cholesterol was notably lower in patients with obstructive CAD ($p<0.05$). Additionally, cardiac chest pain was reported by 82% of the obstructive CAD group, compared with 52.5% in the non-obstructive CAD group, although this difference was not statistically significant ($p>0.05$). A detailed comparison of the baseline characteristics between the two groups is provided in Table 1.

Out of an initial pool of 424 features, we identified the 14 most predictive features for Model 3 using the CatBoost-SHAP approach (Fig. 2C). Of these, the top two features were derived from fat-omics (PQ4_Vol_70_50 and PQ4_Vol_50_30), while the remaining features originated from calcium-omics, highlighting the added value of epicardial fat quantification alongside calcification metrics in predicting obstructive CAD (see Discussion). Interestingly, none of the clinical variables were selected for Model 3, suggesting that imaging-derived features alone carried sufficient predictive power. For comparison, Model 1 retained only four clinical features: age, gender, smoking status, and diabetes mellitus (Fig. 2A). In Model 2, which included clinical and calcium-omics features, 18 calcium-omics variables were selected, while clinical characteristics were excluded (Fig. 2B). Feature importance rankings for all models, as determined through CatBoost and SHAP analysis, are summarized in Supplementary Tables S2-S4.

Model 3 achieved strong predictive performance for obstructive CAD, with a sensitivity of 83.1±4.6%, specificity of 93.8±1.7%, accuracy of 85.3±2.0%, and an F1 score of 73.9±3.3% (Table 2). These results demonstrate a substantial improvement over Models 1 and 2, highlighting the added value of incorporating both calcium-omics and fat-omics features. Notably, the inclusion of fat-omics features led to a marked improvement in model performance, increasing the AUROC from 0.886 to 0.918 and the AUPRC from 0.738 to 0.823, compared to Model 2 (Fig. 3). This improvement emphasizes the relevance of epicardial fat characteristics in refining obstructive CAD prediction. When comparing different machine learning algorithms, CatBoost consistently outperformed the other models (Table 3). While RF and XGBoost yielded nearly equivalent classification results, the LightGBM approach showed the lowest overall performance.

The predictive model demonstrated reliable performance across a wide range of coronary artery calcium (CAC) scores. As illustrated in Fig. 4A, several cases with zero or only minimal-to-mild calcification were correctly identified as having obstructive CAD, whereas examples with high CAC scores were accurately classified as non-obstructive (Fig. 4H). Similarly, cases with moderate CAC scores were appropriately categorized into obstructive and non-obstructive groups, as shown in Fig. 4E and 4F, respectively. These findings highlight the model's ability to integrate both calcification and fat-derived imaging features, reinforcing the added predictive value of fat-omics beyond what is captured by calcium scoring alone.

## 4 Discussion

Building on our prior work in CT-based cardiovascular imaging [7,8,17–22], we developed a novel machine learning framework to predict obstructive CAD using quantitative analysis of coronary calcification and epicardial fat assessments from CTCS scans. This study presents several key contributions to the field. First, we utilized a high-dimensional, biologically meaningful feature set that included 189 calcium-omics and 211 fat-omics features.

These quantitative descriptors provided a detailed characterization of coronary atherosclerosis and epicardial fat distribution, capturing subtle and complementary information that conventional calcium scoring may overlook. Second, our results demonstrated the substantial added value of calcium-omics and fat-omics in predicting obstructive CAD, beyond what is achievable with traditional Agatston scoring alone. Importantly, our model accurately classified extreme cases, including patients with zero or minimal CAC scores who had obstructive disease, and these with high CAC scores but no obstructive disease, highlighting the utility of combining diverse imaging biomarkers to address limitations in standard risk stratification. Third, we performed a systematic comparison of state-of-the-art machine learning algorithms and identified CatBoost as the most effective model for our study, providing superior performance across all key evaluation metrics. Together, these contributions highlight the potential of opportunistically obtained non-contrast CT scans not only for risk stratification but also for proactive disease detection.

We found that volumetric features of epicardial fat tissue, particularly those localized near the coronary arteries, provided meaningful predictive value for obstructive CAD. Among the 14 most predictive features in Model 3, the top two were fat-omics variables, PQ4_Vol_70_50 and PQ4_Vol_50_30, which represent the volume of epicardial fat within HU ranges of -70 to -50 and -50 to -30, respectively, localized to the superior 25% of the heart volume. This region generally corresponds to the anatomical location of the proximal segments of the major coronary arteries, including the left anterior descending, right coronary, and left circumflex arteries. Fat in these HU ranges tends to be less lipid-rich and more attenuated, which has been associated with local inflammation and fibrotic remodeling of pericoronary adipose tissue. Such inflammatory changes in epicardial fat have been strongly linked to vascular inflammation and coronary plaque progression, both of which contribute to the development of obstructive CAD [23–25]. In contrast, epicardial fat with lower attenuation values (e.g., -170 to -150 HU) reflects metabolically inert, lipid-rich fat, which is less indicative of pathological remodeling or inflammatory burden. Vascular inflammation promotes perivascular fibrosis, macrophage infiltration, and proteolytic activity, which in turn may influence vessel wall remodeling and the progression of luminal narrowing. Because these processes are often localized near the coronary arteries and are reflected by elevated HU in surrounding fat tissue, it is likely that fat-omics features in higher HU ranges and proximal regions serve as strong non-invasive markers for obstructive CAD.

The CatBoost model consistently outperformed other widely used machine learning algorithms, including RF, XGBoost, and LightGBM, in predicting obstructive CAD. Several factors contributed to its superior performance in this context. Unlike many gradient boosting frameworks, CatBoost is specifically optimized for structured data and incorporates several innovations that make it highly suitable for complex medical imaging and clinical datasets. First, CatBoost natively supports categorical variables through an efficient encoding strategy that avoids traditional one-hot encoding, thereby preserving information and reducing dimensionality. While our feature set was primarily quantitative, this property still improved model efficiency during internal data handling and could be advantageous in broader clinical applications. More importantly, CatBoost employs an ordered boosting technique that reduces prediction bias by addressing target leakage during training, a limitation common to traditional gradient boosting algorithms like XGBoost and LightGBM. This contributes to more stable and generalizable models, especially in datasets with subtle inter-feature dependencies, such as calcium-omics and fat-omics. Additionally, CatBoost demonstrates strong regularization by design, which helps mitigate overfitting without extensive hyperparameter tuning. Its ability to handle missing values natively and to model complex, nonlinear relationships further improve its robustness. In our study, these features likely enabled CatBoost to capture nuanced patterns across high-dimensional, heterogeneous inputs more effectively than other models. Given these advantages, CatBoost appears to be particularly well-suited for predictive modeling tasks in medical imaging and risk stratification, where data variability and subtle signal patterns are common.

A zero CAC score has traditionally been considered a strong indicator of the absence or minimal presence of CAD and is often used as a binary gatekeeper for further evaluation in symptomatic patients. This concept, commonly referred to as the "power of zero," is based on its high negative predictive value. Numerous studies have validated the low obstructive CAD rates among individuals with zero CAC [26,27]. For instance, Winther et al. [27] reported that only 1.1% of over 25,000 patients with CAC=0 had obstructive CAD on CCTA, yielding a negative predictive value >98%. Similarly, a recent meta-analysis including more than 92,000 patients presenting with stable or acute chest pain demonstrated that a CAC score of zero had a negative predictive value of 97% and 98% for excluding obstructive CAD in stable and acute chest pain populations, respectively [26]. Despite this reassuring profile, CAC=0 does not universally exclude the presence of obstructive CAD, particularly in specific populations. Elevated rates of non-calcified, high-risk plaques have been observed in younger adults (<40 years), women, and patients with diabetes or metabolic disorders [28]. In such groups, reliance on CAC scoring alone may lead to underdiagnosis, as non-calcified plaques are not captured by calcium scoring. Our study addressed this critical limitation by incorporating epicardial fat analysis alongside calcium-omics from non-contrast CTCS scans. By leveraging imaging features beyond calcification alone, our machine learning model successfully classified challenging cases, such as individuals with zero CAC who nevertheless had obstructive CAD confirmed on CCTA. This capability reflects the model's potential to improve diagnostic accuracy where conventional CAC scoring falls short. Given its reliance on widely available non-contrast CT data, our approach may improve the clinical utility of CTCS by enabling more precise, data-driven prediction of obstructive CAD in routine imaging workflows.

This study has several limitations. First, the relatively modest sample size may constrain the generalizability of our findings. Although the results are promising, they should be validated in larger, more diverse populations. Future research involving extensive internal and external datasets will be necessary to assess the model's robustness and applicability across different clinical settings. Second, while our model demonstrated strong predictive performance, questions remain regarding the reproducibility of quantitative features extracted from CTCS images. Specifically, scan-to-scan variability in image acquisition or reconstruction could affect consistency. To address this, future studies should incorporate repeat imaging (i.e., scan-rescan protocols) to evaluate reproducibility and measurement stability. Third, although the calcium-omics and fat-omics extraction processes are fully automated, the total processing time per scan remains approximately 10-15 minutes, which may hinder large-scale clinical implementation. Continued efforts in software optimization and integration into clinical workflows will be essential to improve efficiency and support broader adoption.

# 5 Conclusion

In this study, we developed and validated a novel machine learning framework for the prediction of obstructive CAD using quantitative assessments of coronary calcification and epicardial fat derived from non-contrast CTCS scans. By integrating calcium-omics and fat-omics features, our model successfully identified patients with obstructive CAD, including those with challenging presentations such as zero CAC scores. These results highlight the potential of CTCS as a low-cost, non-invasive tool not only for cardiovascular risk stratification but also for direct disease detection. Our approach may help inform clinical decision-making and reduce reliance on contrast-enhanced or invasive testing, particularly in low- to intermediate-risk populations.

## Funding Declaration

This project was supported by the National Heart, Lung, and Blood Institute through grants K01HL171795, R01HL167199, and NIH R01HL165218. This research was conducted in space renovated using funds from an NIH construction grant (C06 RR12463) awarded to Case Western Reserve University.

## Acknowledgments

The content of this report is solely the responsibility of the authors and does not necessarily represent the official views of the National Institutes of Health. The grants were obtained via collaboration between Case Western Reserve University and University Hospitals of Cleveland. This work made use of the High-Performance Computing Resource in the Core Facility for Advanced Research Computing at Case Western Reserve University. The veracity guarantor, Justin N. Kim, affirms to the best of his knowledge that all aspects of this paper are accurate.

## Competing interests

The authors declare no competing interests.

## Data availability statement

The data that support the findings of this study are available from University of Edinburgh, but restrictions apply to the availability of these data, which were used under license for the current study, and so are not publicly available. Data are, however, available from the authors upon reasonable request and with permission of the University of Edinburgh.

## Author contributions

M.H.E.M contributed to the ideation, overview, and drafting of the manuscript. A.H. and T.H. contributed to the formal analysis and data preparation. H.W. and J.N.K. contributed to the study design and analysis. M.C.W. and D.E.N contributed to data acquisition and interpretation. R.G., S.R., and D.L.W. contributed to the conception and design of the work. J.L. contributed to drafting of the manuscript, review, editing, supervision, and funding acquisition. All authors read and approved the final manuscript.

## Tables

Table 1 Baseline clinical characteristics of patients with obstructive (n=334) and non-obstructive CAD (n=990) (BMI: body mass index, HDL: high-density lipoprotein, CHD: congenital heart disease, and ACE: (angiotensin-converting enzyme).

| Features | Total<br>(*n*=1,324) | Obstructive<br>(*n*=334) | Non-obstructive<br>(*n*=990) | *p*-value |
|---|---|---|---|---|
| Male | 737/1,324 (55.7%) | 254/334 (76.0%) | 483/990 (48.8%) | 0.418 |
| BMI | 29.6±5.6 | 29.4±4.8 | 29.7±5.8 | 0.513 |
| Age | 57.3±9.4 | 61.2±7.7 | 55.9±9.5 | <0.00001 |
| BMI band (≥30 or <30) | 534/1,324 | 132/334 (39.5%) | 402/990 (40.6%) | 0.515 |
| Age band (18-59 or 60-75) | 577/1,324 | 196/334 (58.7%) | 381/990 (38.5%) | 0.670 |
| Diabetes Mellitus | 148/1,324 | 45/334 (13.5%) | 103/990 (10.4%) | 0.875 |
| Height | 1.70±0.10 | 1.72±0.1 | 1.69±0.10 | <0.001 |
| Weight | 85.4±17.8 | 86.7±16.1 | 84.9±18.3 | 0.111 |
| Smoking habit* | Non: 631/1323 (47.7%)<br>Prior: 441/1323 (33.3%)<br>Current: 251/1323 (19.0%) | Non: 149/334 (44.6%)<br>Prior: 119/334 (35.6%)<br>Current: 66/334 (19.8%) | Non: 482/989 (48.7%)<br>Prior: 322/989 (32.6%)<br>Current: 185/989 (18.7%) | 0.729 |
| Cigarettes per day | 2.8±7.1 | 2.9±6.8 | 2.8±7.2 | 0.949 |
| Hypertension | 459/1,324 (34.7%) | 151/329 (45.9%) | 308/981 (31.4%) | 0.556 |
| Total cholesterol | 5.0±1.9 | 5.1±2.0 | 5.0±1.8 | 0.545 |
| HDL cholesterol | 1.0±0.7 | 0.9±0.6 | 1.1±0.7 | <0.01 |
| CHD family history | 584/1,324 (44.1%) | 143/333 (42.9%) | 441/981 (45.0%) | 0.946 |
| Systolic blood pressure | 138.3±24.0 | 142.2±21.6 | 137.0±24.6 | <0.001 |
| Diastolic Blood Pressure | 81.2±13.7 | 82.1±12.5 | 81.0±14.1 | 0.195 |
| Chest pain<br>(cardiac vs. non-cardiac) | 794/1,324 (60.0%) | 274/334 (82.0%) | 520/990 (52.5%) | 0.685 |
| Antiplatelet | 801/1,324 (60.5%) | 303/334 (90.7%) | 498/990 (50.3%) | 0.191 |
| Statin | 737/1,324 (55.7%) | 286/334 (85.6%) | 451/990 (45.6%) | 0.914 |
| ACE inhibitor | 230/1,324 (17.4%) | 96/334 (28.7%) | 134/990 (13.5%) | 0.567 |
| Calcium blocker | 125/1,324 (9.4%) | 48/334 (14.4%) | 77/990 (7.8%) | <0.01 |
| Nitrates | 335/1,324 (25.3%) | 140/334 (41.9%) | 195/990 (19.7%) | 0.920 |
| Betablocker | 463/1,324 (35.0%) | 216/334 (64.7%) | 247/990 (24.9%) | 0.180 |
| Hyperlipidemia | 802/1,324 (60.6%) | 277/334 (82.9%) | 525/990 (53.0%) | 0.722 |

*The total is 1,323 due to one missing data point.

Table 2 Classification performance of the three models using 1,000 repeated 5-fold cross-validation. Identical training and testing splits were used for all models, with hyperparameter tuning via grid search. Model 3 demonstrated superior performance, highlighting the added predictive value of calcium-omics and fat-omics features.

| **Models (CatBoost)** | **Sensitivity** | **Specificity** | **Accuracy** | **F1** |
|---|---|---|---|---|
| Model 1 (Clinical only) | 69.6±7.6 | 86.8±2.6 | 67.6±3.1 | 51.9±4.9 |
| Model 2 (Clinical + Calcium-omics) | 80.9±5.7 | 92.7±2.2 | 81.5±2.2 | 68.6±4.1 |
| Model 3 (Clinical + Calcium-omics + Fat-omics) | 83.1±4.6 | 93.8±1.7 | 85.3±2.0 | 73.9±3.3 |

Table 3 Comparison of machine learning classifiers trained with combined clinical, calcium-omics, and fat-omics features. Results for RF, XGBoost, LightGBM, and CatBoost were reported based on 1,000 repeated 5-fold cross-validation. The same feature selection strategy and training/testing splits were used across all models, with grid search for hyperparameter optimization.

| **Models** | **Sensitivity** | **Specificity** | **Accuracy** | **F1** |
|---|---|---|---|---|
| RF | 70.4±5.4 | 90.3±1.6 | 87.0±1.7 | 73.2±3.7 |
| XGBoost | 70.1±5.1 | 90.2±2.4 | 87.2±2.1 | 73.3±4.0 |
| LightGBM | 38.0±38.9 | 83.3±9.1 | 80.6±6.5 | 73.6±6.3 |
| CatBoost | 83.1±4.6 | 93.8±1.7 | 85.3±2.0 | 73.9±3.3 |

## Figures with Figure Legends

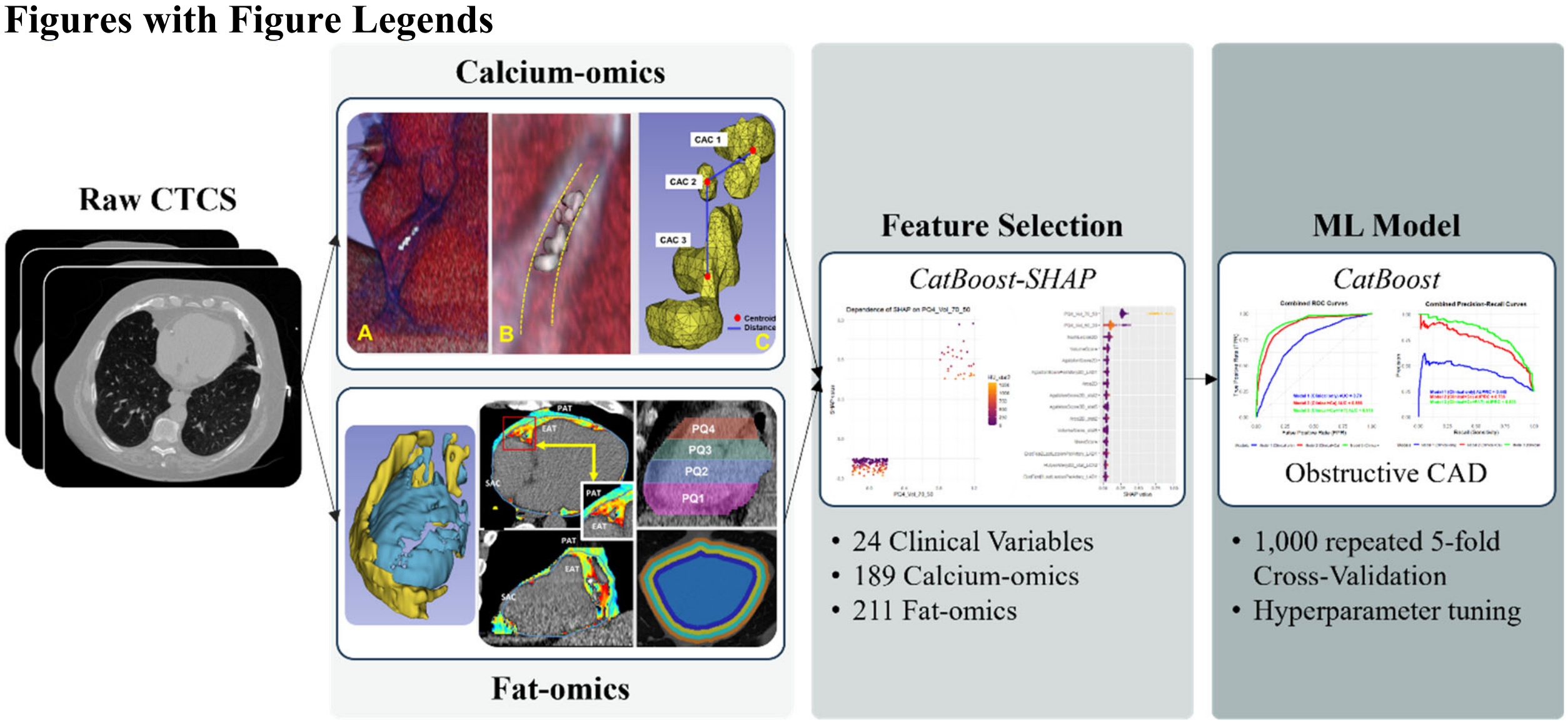

Fig. 1 Workflow for predicting obstructive CAD using non-contrast CTCS scans. The pipeline consists of three key steps: feature extraction, feature selection, and machine learning-based prediction. A total of 424 features were extracted, including 24 clinical variables, 189 calcium-omics features, and 211 fat-omics features. The most predictive features were selected using CatBoost-SHAP analysis and used to train a CatBoost model for obstructive CAD prediction.

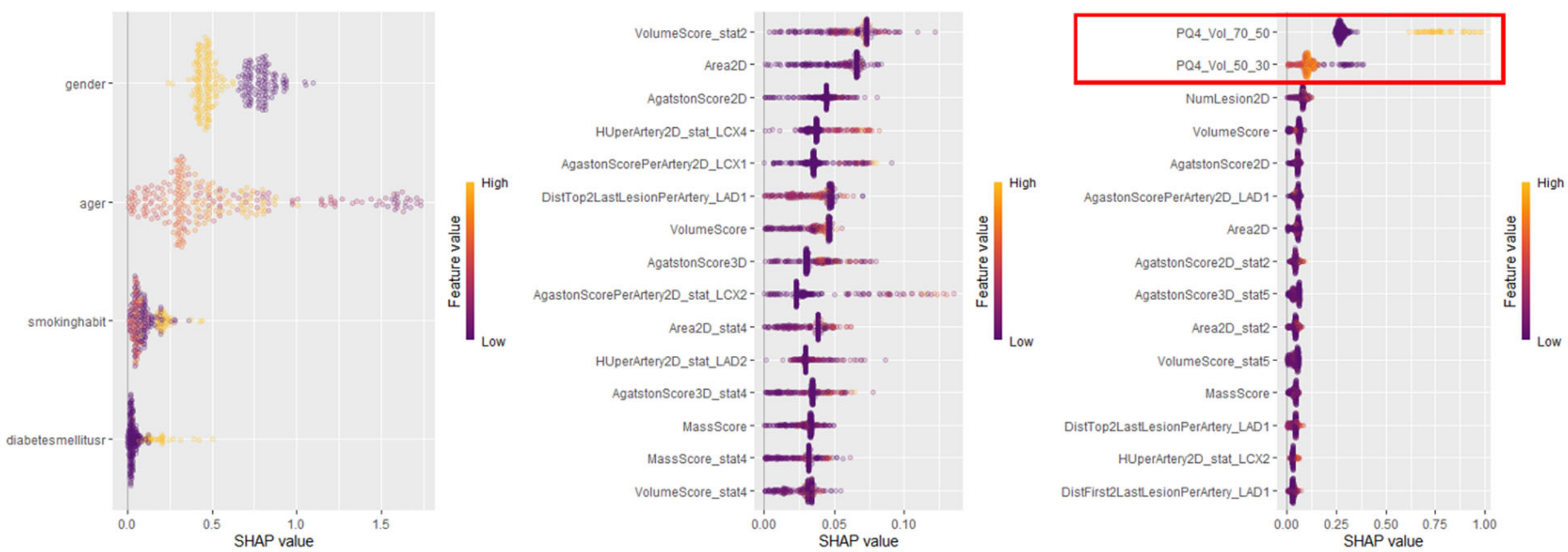

Fig. 2 Feature importance plots of the three predictive models. Left: Model 1 (clinical features only); Center: Model 2 (clinical + calcium-omics); Right: Model 3 (clinical + calcium-omics + fat-omics). Feature selection was performed using CatBoost and SHAP analysis, as described in Section 2.5. In Model 3, 14 features were selected, with the top two derived from fat-omics (PQ4_Vol_70_50 and PQ4_Vol_50_30, highlighted in the red box).

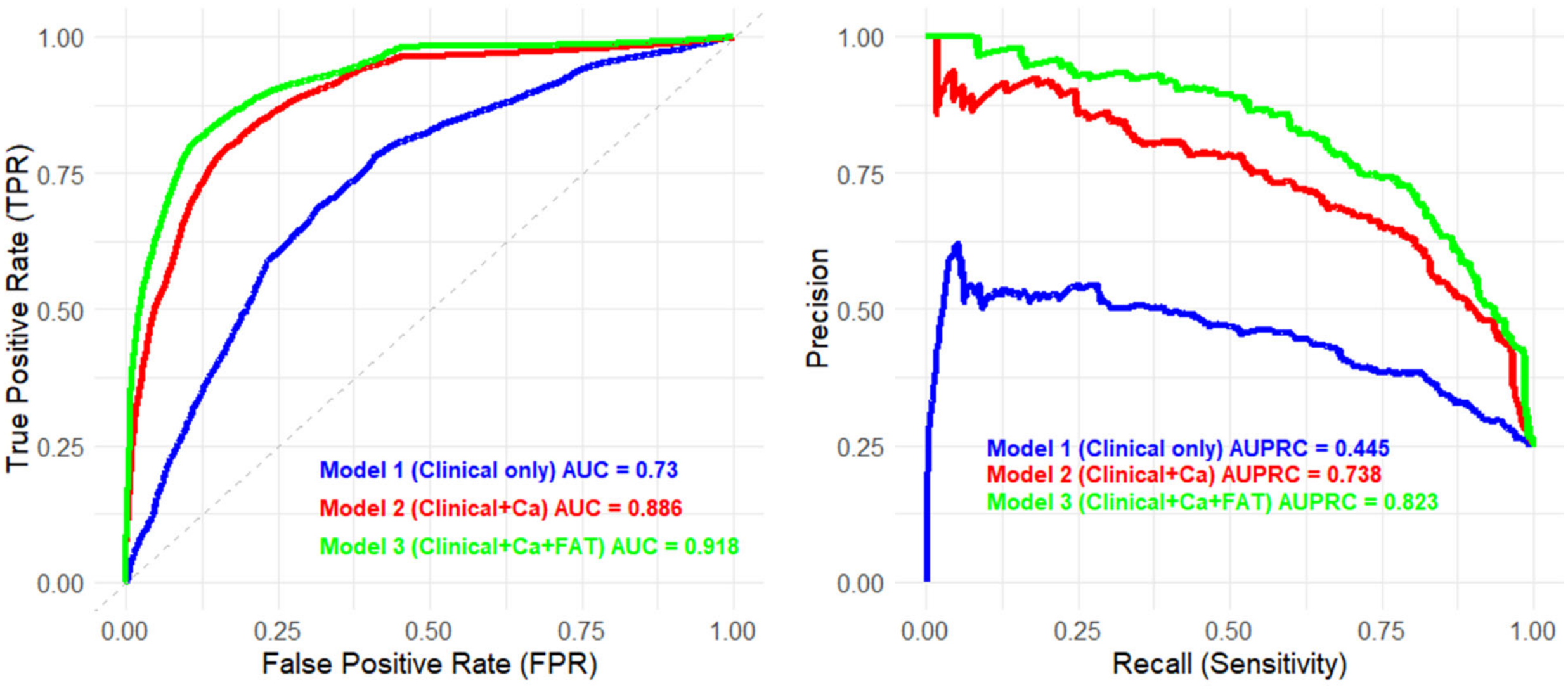

Fig. 3 AUROC (left) and AUPRC (right) curves for each model. Model 3 consistently achieved the highest classification performance, with AUROC increasing from 0.730 to 0.918 and AUPRC from 0.445 to 0.823 when including calcium-omics and fat-omics.

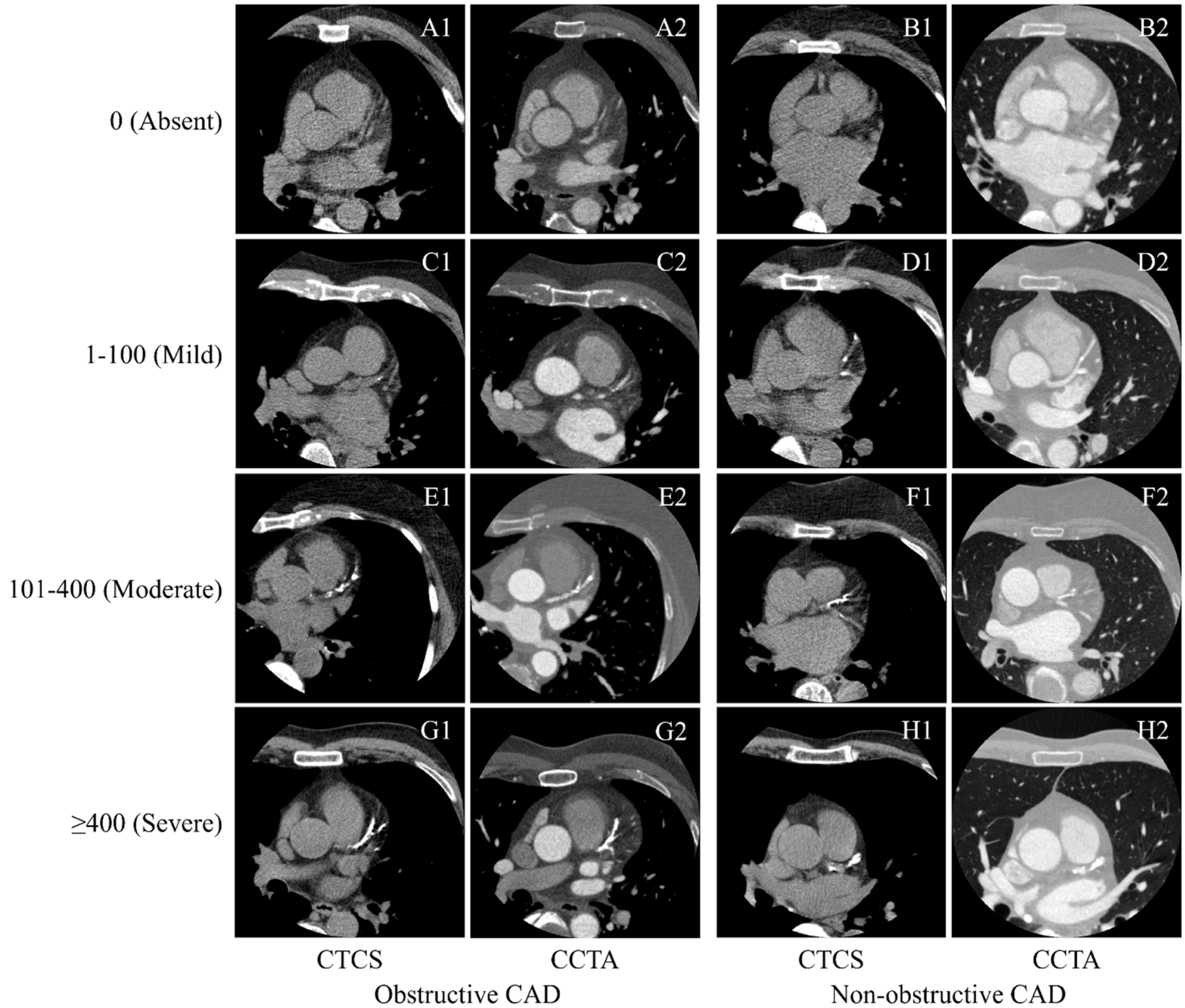

Fig. 4 Representative paired CTCS and CCTA images across CAC score categories in patients with and without obstructive CAD. Each row corresponds to a different CAC category: absent (0), mild (1–100), moderate (101–400), and severe (≥400). The left and right columns show CTCS and corresponding CCTA images, respectively, for patients with obstructive (left column) and non-obstructive (right column) CAD. Notably, in panels A1 and A2, a patient with zero CAC was correctly classified as having obstructive CAD, while in H1 and H2, a patient with severe CAC was accurately classified as non-obstructive. These examples highlight the model's ability to leverage epicardial fat features beyond traditional calcification measures to improve diagnostic accuracy.